\documentclass[sigconf]{acmart}

\usepackage{multirow} 
\usepackage{graphicx}  
\usepackage{caption}
\usepackage{algorithm}      
\usepackage{algorithmic}  
\usepackage[table]{xcolor}
\usepackage{tabularray}
\usepackage{makecell}

\newtheorem{theorem}{Theorem}
\newtheorem{lemma}{Lemma}
\newtheorem{assumption}{Assumption}

\definecolor{lightgray}{rgb}{0.85,0.85,0.85}
\definecolor{lightorange}{rgb}{1, 0.85, 0.8}

\AtBeginDocument{%
  }

\setcopyright{acmlicensed}
\copyrightyear{2026}
\acmYear{2026}
\setcopyright{cc}
\setcctype{by}
\acmDOI{10.1145/3767308.3835033}
\acmConference[MM '26]{Proceedings of the 34th ACM International Conference on Multimedia}{November 10--14, 2026}{Rio de Janeiro, Brazil.}
\acmISBN{979-8-4007-2213-4/2026/11}

\begin{document}

\title[TOFD]{TOFD: Target-Oriented Feature Decoupling against \\Poisoning Attacks in Split Federated Learning}


\author{Yuhan Xie}
\authornote{Both authors contributed equally to this research.}
\orcid{https://orcid.org/0009-0007-3301-0888}
\affiliation{%
  \institution{Shanghai University of Finance and Economics}
 \department{MoE Key Laboratory of Interdisciplinary Research of Computation and Economics}
  \city{Shanghai}
  \country{China}
}
\email{yhtse@stu.sufe.edu.cn}

\author{Jingrong Huang}
\authornotemark[1]
\orcid{https://orcid.org/0009-0009-4694-1344}
\affiliation{%
  \institution{Shanghai University of Finance and Economics}
  \city{Shanghai}
  \country{China}
}
\email{jingrhuang@stu.sufe.edu.cn}

\author{Chen Lyu}
\correspondingauthor
\authornote{Corresponding author.}
\orcid{https://orcid.org/0000-0002-4373-9898}
\affiliation{%
  \institution{Shanghai University of Finance and Economics}
  \department{MoE Key Laboratory of Interdisciplinary Research of Computation and Economics}
  \city{Shanghai}
  \country{China}
}
\email{lyu.chen@mail.shufe.edu.cn}

\renewcommand{\shortauthors}{Yuhan Xie, Jingrong Huang, \& Chen Lyu}


\begin{abstract}
    Split Federated Learning (SFL) facilitates privacy-preserving collaborative training with reduced client-side overhead. However, its split architecture introduces unique attack surfaces, rendering it vulnerable to diverse poisoning attacks. Most existing defenses fail to exploit the split paradigm, limiting their ability to detect and contain malicious behaviors at an early stage. To bridge this gap, we propose Target-Oriented Feature Decoupling (TOFD), a unified framework that jointly enables proactive detection and robust optimization against a wide range of poisoning attacks. TOFD operates in three stages: (1) Target Inference, which identifies potential attack targets by refining class-wise safe zones via class-specific Margin Perturbation (MP); (2) Sample Purification, which adaptively filters poisoned smashed data using thresholds calibrated through cross-class min–max normalization of MP; and (3) Decoupling Optimization, which leverages an adversarial guidance model to capture attack-induced patterns and decouple their influence during optimization, thereby suppressing residual adversarial effects. We provide theoretical guarantees for the convergence of TOFD. Extensive experiments on five datasets demonstrate that TOFD consistently outperforms state-of-the-art defenses under diverse attack scenarios, achieving superior robustness with low computational overhead suitable for practical deployment.
\end{abstract}




\maketitle

\section{Introduction}
\begin{figure}[t]
	\centering   \includegraphics[width=1\linewidth]{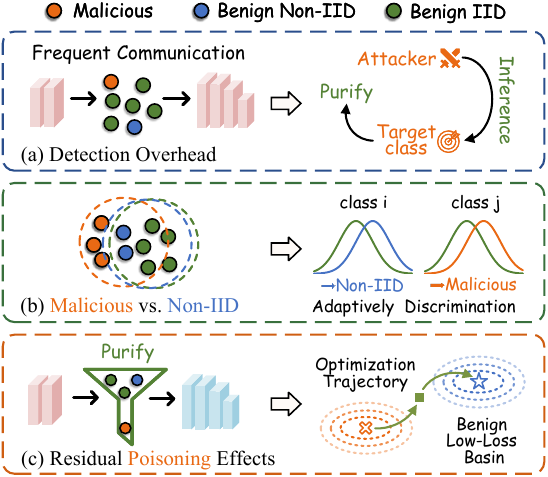}
    \vspace{-4mm}
	\caption{Motivation: Illustration of poisoning defense challenges in SFL and the intuition behind TOFD. }
	\label{fig:motivation}
\end{figure}
Split Federated Learning (SFL) \cite{thapa2022splitfed,li2024introducing} enables collaborative training on decentralized data by combining the strengths of Federated Learning (FL) \cite{yazdinejad2024robust} and Split Learning (SL) \cite{lin2024efficient}. This paradigm preserves data privacy while significantly reducing the computational burden on resource-constrained clients. Specifically, SFL partitions the global model into client and server components. Clients execute the initial forward pass and transmit intermediate representations, known as smashed data, to the server. The server then completes the remaining computations and propagates gradients back to facilitate local updates. Finally, a dual-aggregation mechanism is employed to coordinate the system: the fed server aggregates client-side components, while the main server aggregates server-side components.

Despite its promise as a computationally efficient and privacy-preserving paradigm \cite{xie2026besplit}, SFL is inherently susceptible to diverse poisoning attacks \cite{sandeepa2024sherpa,wu2024evaluating}. This vulnerability primarily stems from its partitioned and collaborative training architecture, exposing multiple attack surfaces throughout the learning pipeline \cite{xie2026healsplit}. Hence, by exploiting these attack surfaces, attackers can strategically manipulate various segments of the training process, including local feature representations \cite{li2025infighting}, labels \cite{jha2023label,jiangfedclean}, smashed data \cite{wu2024evaluating}, and client-side model weights \cite{fang2020local,yazdinejad2024robust}. Such multifaceted interventions facilitate the injection of destructive signals into the training process, thereby undermining aggregation and ultimately degrading global model performance.

To defend against poisoning attacks in SFL, most existing countermeasures 
\cite{yin2018byzantine,guerraoui2018hidden,cao2020fltrust,wu2024evaluating,li2024data,kumar2025fortifying} primarily adapt defense strategies originally designed for conventional FL. However, these methods offer limited efficacy in practice, as they fail to exploit SFL's partitioned architecture for early-stage intervention. This limitation largely arises from overlooking the pivotal role of smashed data \cite{xie2026besplit}. Unlike standard FL, server-side optimization in SFL is explicitly dependent on client-transmitted smashed data, which serves as the primary conduit through which poisoning signals propagate across training stages \cite{li2024introducing}. We argue that this dependency, while increasing vulnerability, simultaneously introduces a unique defensive opportunity.  Since diverse adversarial perturbations from any source tend to manifest as anomalies in this intermediate representation, intercepting these signals prior to server-side computation establishes a strategic checkpoint capable of alleviating a wide spectrum of poisoning attacks. This insight naturally raises a fundamental question: how to reliably identify poisoned smashed data and prevent it from compromising server-side optimization in SFL?

In practice, effectively realizing this defensive opportunity in SFL faces several challenges.
First, the training process involves frequent transmission of smashed data, which incurs substantial communication and detection overhead, thereby necessitating lightweight and low-cost defense mechanisms \cite{wu2024evaluating}.
Second, in dynamic SFL systems, the presence of benign non-IID clients alongside malicious ones leads to complex distributional shifts, making it difficult to distinguish non-IID behavior from adversarial manipulation and increasing the risk of misdetection \cite{krauss2023mesas,xie2024fedredefense}.
Third, even when poisoned smashed data are filtered, malicious client-side models may still contaminate SFL optimization through aggregation, necessitating system-level mitigation of residual adversarial effects. Based on the outlined challenges, Figure~\ref{fig:motivation} illustrates the corresponding design insights that guided our framework development.

% To overcome the outlined challenges, 
Building upon these insights, we propose \textbf{T}arget-\textbf{O}riented \textbf{F}eature \textbf{D}ecoupling (TOFD), a unified defense framework that synergizes early-stage detection with server-side optimization to counter a broad spectrum of poisoning attacks in SFL. TOFD is built upon three complementary pillars. First, TOFD leverages class-wise inference to localize potential attack targets, thereby enabling low-overhead detection. By narrowing the scope to suspicious classes, the framework establishes  client-level safe zones, which are further refined via Margin Perturbation (MP) to effectively disentangle adversarial behavior from benign non-IID variation. Second, TOFD adopts fine-grained, class-aware sample filtering to mitigate poisoning effects while preserving data diversity. Rather than discarding entire clients, the framework selectively prunes anomalous samples within identified target classes using thresholds calibrated via min–max normalization of the MP \cite{arp2022and}. Consequently, only retained benign representations are utilized to update global distributions and optimize the server-side model. Third, to address residual adversarial influence that may persist beyond detection, TOFD integrates a decoupling objective into the SFL optimization. Specifically, an adversarial guidance model is developed to capture attack-induced patterns from malicious data, allowing the server to suppress adversarial representations while preserving benign feature integrity. Once converged, the adversarial guidance model serves as a stable, plug-and-play component for SFL optimization. Our main contributions can be summarized as follows:

\begin{itemize}
    \item We propose TOFD, the first framework to systematically integrate fine-grained detection with SFL optimization, enabling effective mitigation of a wide range of poisoning attacks, including data, weight, smashed, label, and multi-vector poisoning.
       
    \item   By leveraging class-wise inference and MP, TOFD disentangles malicious behavior from benign non-IID variations while preserving data diversity. Furthermore, an adversarial guidance–based decoupling strategy is developed to suppress residual attack influence.
        
    \item We provide a formal theoretical analysis of TOFD, characterizing its computational complexity and establishing convergence guarantees.

    \item Extensive experiments across five benchmarks and diverse model architectures demonstrate that TOFD consistently outperforms state-of-the-art defenses in both robustness and efficiency, even under extreme data heterogeneity.
\end{itemize}

\begin{figure*}[t]
	\centering
	\includegraphics[width=1\textwidth]{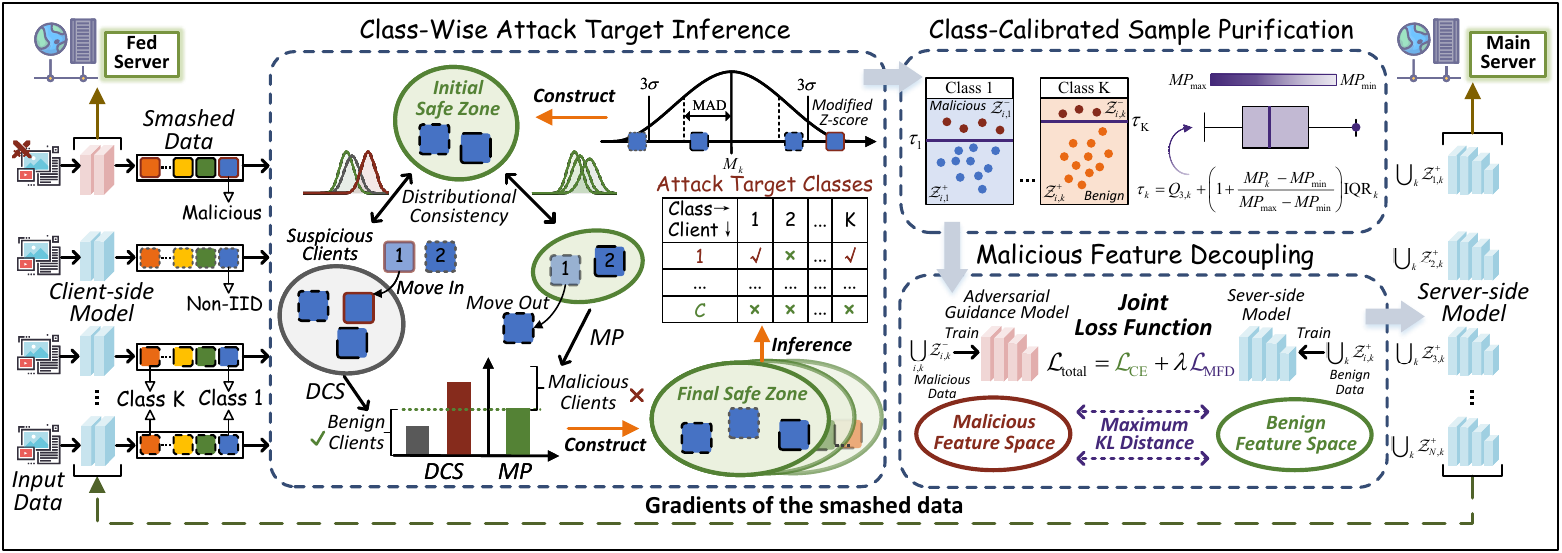}
    \vspace{-4mm}
	\caption{The framework of TOFD. After the smashed data are transmitted to the server, TOFD first constructs an initial class-wise safe zone using a modified Z-score and then refines it by evaluating the Distributional Consistency Score (DCS) against the MP threshold.
		Clients that fall outside the safe zone are inferred to be attacking the corresponding class.
		For the identified target classes, poisoned samples are filtered using an adaptive threshold $\tau_k$, yielding validated benign sets $\mathbf{Z}_{i,k}^{+}$ and detected malicious sets $\mathbf{Z}_{i,k}^{-}$ for each client $c_i$. 
		An adversarial guidance model trained on $ \bigcup_{i,k} \mathbf{Z}^-_{i,k}$ is incorporated into the SFL system to decouple  residual adversarial influence through a joint loss. Finally, the updated client- and server-side models are aggregated by the fed server and main server, respectively, completing the defense and optimization process.}
	\label{fig:TOFD}
\end{figure*}

\section{Related Work} \label{Related}

\subsection{Defenses in Split Federated Learning}

Existing defenses for SFL can be broadly categorized into model validation and data validation approaches. Model validation strategies mitigate adversarial influence by scrutinizing the integrity of model updates. Classical Byzantine-resilient methods, such as Krum \cite{blanchard2017machine}, Trimmed Mean \cite{yin2018byzantine}, and Bulyan \cite{guerraoui2018hidden}, employ geometric or order-based statistics to suppress outlier updates. Building on these, FLTrust \cite{cao2020fltrust} incorporates a small trusted dataset on the server side to assign trust scores and normalize update magnitudes, thereby bounding the impact of malicious contributions. In contrast, data validation strategies seek to evaluate the reliability of client data. FedBary \cite{li2024data} treats client valuation as a Wasserstein barycenter problem, quantifying distributional discrepancies to enable contribution-aware aggregation. Similarly, FAVD \cite{kumar2025fortifying} implements a privacy-preserving mechanism to assess client-side data quality through density comparisons.
  
However, most existing approaches remain extensions of conventional FL and are not explicitly tailored to the SFL architecture, where split-model optimization introduces unique and vulnerable attack surfaces \cite{wu2024evaluating}.
While HealSplit \cite{xie2026healsplit} represents a pioneering effort specifically designed for SFL via multi-teacher adversarial distillation, its reliance on generative modeling and exhaustive inspection incurs substantial computational overhead.
This limitation underscores the need for more efficient and architecture-aware defense mechanisms in SFL.

\subsection{Defenses in FL under Data Heterogeneity}
Defending FL under data heterogeneity requires a delicate balance between filtering malicious updates and preserving the benign contributions of diverse clients. Recent studies address this challenge through various perspectives. PRFL \cite{yuan2025prfl} introduces a defense-oriented personalized framework that suppresses malicious participants via similarity-based trust modeling and adaptive knowledge sharing. FedREDefense \cite{xie2024fedredefense} leverages distilled local knowledge to distinguish genuine training dynamics from poisoning-induced deviations. Furthermore, FDCR \cite{huang2024parameter} enhances backdoor robustness in heterogeneous settings by detecting malicious clients through parameter importance and gradient discrepancies, subsequently rescaling critical updates during aggregation.

Despite their effectiveness, these defenses are often specialized for individual attack types (e.g., either data or model poisoning) and struggle to maintain efficacy in more complex adversarial scenarios involving composite poisoning attacks. 

\section{Background}\label{background}
\subsection{Problem Setup}

Let the SFL system consist of $N$ clients $\mathcal{C} = \{c_i\}_{i=1}^N$, comprising benign clients $\mathcal{C}_{\mathrm{ben}}$ and malicious clients $\mathcal{C}_{\mathrm{att}} = \mathcal{C} \setminus \mathcal{C}_{\mathrm{ben}}$. Each client $c_i$ holds a private dataset 
%\( \mathcal{D}_i = \{(x_j, y_j)\}_{j=1}^{m_i^{\mathrm{tr}}} \cup  \{(x_j, y_j)\}_{j=1}^{m_i^{\mathrm{te}}} \sim \mathcal{P}_i \) 
\( \mathcal{D}_i = \mathcal{D}_i^{\mathrm{tr}}  \cup \mathcal{D}_i^{\mathrm{te}} \sim \mathcal{P}_i\) 
containing $K$ classes, where $\mathcal{D}_i^{\mathrm{tr}}$ and $\mathcal{D}_i^{\mathrm{te}}$ denote local training and test sets, and $\mathcal{P}_i$ represents the local data distribution.

At each communication round $t$, the client-side model $g_{\theta_{c_i}}$ maps a local mini-batch of inputs $\{ x_j \}_{j=1}^B$ into smashed data $\{ z_j \}_{j=1}^{B}$, which are transmitted with their corresponding labels $\{y_j\}_{j=1}^{B}$ to the server-side model $h_{\theta_{s_i}}$ for forward computation.
The resulting gradients are back-propagated to update the client-side parameters $\theta_{c_i}$. Subsequently, client-side parameters $\{\theta_{c_i}\}_{i=1}^N$ and server-side parameters $\{\theta_{s_i}\}_{i=1}^N$ are aggregated by the fed server and the main server, respectively, yielding the global model 
$f_{\theta} = g_{\theta_c} \circ h_{\theta_s} $.

\textbf{Defender Objective.}  The first objective is to accurately identify malicious smashed data, measured by the Malicious Sample Detection Rate (MSDR):
\begin{equation}
	\mathrm{MSDR} = \frac{1}{|\mathcal{M}|} \sum_{z \in \mathcal{M}} \mathbb{I}_{mal}(z),
\end{equation}
where $\mathcal{M}$ denotes the set of malicious smashed data, and $\mathbb{I}_{mal}(z)$ is an indicator function.

The second objective is to achieve robustness against diverse poisoning attacks while maintaining clean performance:%formulated as an optimization problem:
\begin{equation}
	\min_{\theta} \mathbb{E}_{(x,y) \sim \mathcal{D}^{te}} \left[ \ell(f_\theta(x), y) \right] + \mu \mathcal{R}_{\text{robust}}(\theta),
\end{equation}
where \( \ell(\cdot, \cdot) \) is a task-specific loss function, \( \mathcal{R}_{\text{robust}}(\theta) \) measures the sensitivity to attacks, and $\mu$ weights the robustness regularization.

\subsection{Threat Model} 

In SFL, malicious clients with heterogeneous knowledge may inject perturbations at different stages of the training workflow, giving rise to a broad spectrum of poisoning threats, which are categorized into five representative attack types:

\textbf{Data Poisoning (DP, ${O}_{1}$):} 
The local dataset is manipulated as   
\( \mathcal{D}_k' = \{(x_j + \delta_x, y_j)\}_{j=1}^{m'} \),  
where \( \delta_x \) is the input perturbation.

\textbf{Weight Poisoning (WP, ${O}_{2}$):} The model parameters are manipulated before aggregation as 
\( \theta' = \theta + \Delta_\theta \), where \( \Delta_\theta \) represents the weight perturbation.

\textbf{Smashed Poisoning (SP, ${O}_{3}$):} 
The smashed dataset is manipulated as
\( \mathbf{Z}_k' = \{ z_j + \delta_z \}_{j=1}^{m'} \),
where $\delta_z$  is the applied perturbation.

\textbf{Label Poisoning (LP, ${O}_{4}$):} 
The label set is manipulated as
\( \mathcal{Y}  _k' = \{ (y_j + \delta_y) \bmod K \}_{j=1}^{m'} \),  
where \( \delta_y \) is  the label shift.

\textbf{Multi-Vector Poisoning:}
Multiple attack strategies may be applied simultaneously:
${O}_{\text{multi}} = \bigcup_{i=1}^{4} \mathbb{I}_i {O}_i,
\, \mathbb{I}_i \in \{0,1\}$,
where \( \mathbb{I}_i \) indicates whether the \( i \)-th attack is applied.

\section{Methodology}
In this section, we introduce the details of TOFD for defending SFL. 
TOFD integrates detection and optimization in a three-stage pipeline: it first infers the attack target classes of malicious clients, then filters poisoned samples within the inferred classes, and finally suppresses residual adversarial influence. 
Figure~\ref{fig:TOFD} illustrates the overall framework.

\subsection{Class-Wise Attack Target Inference}

\begin{figure*}[t]
    \centering
    \begin{minipage}[t]{1\textwidth}
        \captionof{table}{Comprehensive overview of the default experimental setup.}
            \vspace{-2mm}
        \label{tab:setting}
        \resizebox{\textwidth}{!}{%
            \begin{tabular}{lcccccccc}
                \hline
                \rowcolor{gray!15}
                \textbf{Dataset} &
                \makecell{\textbf{Total}\\\textbf{Clients}} &
                \makecell{\textbf{Non-IID}\\\textbf{parameter $\kappa$}} &
                \textbf{Model} &
                \makecell{\textbf{Total}\\\textbf{Epochs}} &
                \makecell{\textbf{Batch}\\\textbf{Size}} &
                \textbf{Attack Settings} &
                \textbf{Other Settings} \\
                \hline
                \hline
                HAM10k~\cite{tschandl2018ham10000} & 64 &
                \multirow{5}{*}{\makecell{0.1, 0.5, 0.3,\\ 1 (Default),\\ 5}} &
                DenseNet121 & 200 & 64 &
                \multirow{5}{*}{\makecell{Malicious ratio = 20\%,\\Coordinate updates = $3\times10^{4}$,\\Attack learning rate = $1\times10^{-2}$,\\Image update rate = $1\times10^{4}$}} &
                \multirow{5}{*}{\makecell{SFL learning rate = $1\times10^{-4}$,\\ $\rho = 0.6745$, Dataset (MNIST),\\ $\lambda,\beta = 0.2$, Attack method (DP+SP),\\Adversarial guidance model\\ (Server-side model)}} \\
                \cline{1-2} \cline{4-6}
                MNIST~\cite{lecun1998mnist} & 100 & & ResNet18 & 300 & 128 & & \\
                \cline{1-2} \cline{4-6}
                F-MNIST~\cite{xiao2017fashion} & 100 & & ResNet18 & 300 & 128 & & \\
                \cline{1-2} \cline{4-6}
                CIFAR10~\cite{krizhevsky2009learning} & 100 & & ResNet18 & 300 & 128 & & \\
                \cline{1-2} \cline{4-6}
                CIFAR100~\cite{krizhevsky2009learning} & 64 & & ResNet50 & 200 & 128 & & \\
                \hline
            \end{tabular}
        }
    \end{minipage}
\end{figure*}

At communication round $t$, each client $c_i$ transmits a batch of smashed data, where the subset corresponding to class $k$ is denoted by $\mathcal{B}_{i,k}^{(t)}$.

\begin{definition}[Empirical Gaussian Modeling]
	\label{def:gauss_op}
	Given a smashed dataset $\mathbf{Z} = \{z_j\}_{j=1}^n$, the empirical Gaussian modeling operator $\mathcal{G}(\cdot)$ maps $\mathbf{Z}$ to a diagonal Gaussian distribution $\mathcal{N}(\mu, \Sigma)$, whose parameters are estimated as
	\begin{equation*}
		\mu = \frac{1}{n} \sum_{j=1}^n \phi(z_j),   \Sigma = \mathbf{D} \Bigl( \frac{1}{n} \sum_{j=1}^n (\phi(z_j) - \mu)(\phi(z_j) - \mu)^\top \Bigr),
	\end{equation*}
	where $\phi(\cdot)$ maps samples into a low-dimensional space and $\mathbf{D}(\cdot)$ extracts the diagonal elements.
\end{definition}

According to Definition~\ref{def:gauss_op},  the class-\(k\) feature distribution of client \(c_i\) is given by
$\mathcal{P}_{i,k}^{(t)} = \mathcal{G}(  \mathcal{B}_{i,k}^{(t)} )$.

\subsubsection{Initial Class-Wise Safe Zone Construction}

To establish a class-wise safe zone for benign clients, we first measure the distance between the local distribution $\mathcal{P}_{i,k}^{(t)}$ and the historical global distribution $\mathcal{P}_{G,k}^{(t-1)}$, which is maintained using identified clean smashed data:
\begin{equation}
	\label{eq:dist_gaussians}
	\mathcal{W}_{i,k}^{(t)} = {dist}\left(
	\mathcal{P}_{i,k}^{(t)},\,
	\mathcal{P}_{G,k}^{(t-1)}
	\right).
\end{equation}
where ${dist}(\cdot,\cdot)$ denotes the 2-Wasserstein distance.

Based on these distances, the initial safe zone $\mathcal{C}_{k}^{(t)}$, consisting of high-confidence benign clients, is constructed via the modified Z-score:
%\cite{ICLR2024_modifiedz}:
\begin{equation}
	\label{eq:init_safe}
	\mathcal{C}_{k}^{(t)} = \{ c_i \in \mathcal{C} :  \frac{ \mathcal{W}_{i,k}^{(t)} - M_k^{(t)}} {\text{MAD}_k^{(t)}/\rho} \le 3\sigma \}.
\end{equation}
where $\rho$ is a constant set to 0.6745, as recommended in \cite{zscore}, $M_k^{(t)}$ and $\text{MAD}_k^{(t)}$ denote the median and median absolute deviation of $\{\mathcal{W}_{i,k}^{(t)}\}_{c_i \in \mathcal{C}}$, respectively. 
Clients that fall outside $\mathcal{C}_{k}^{(t)}$ form the set of suspicious clients, denoted as $\tilde{\mathcal{C}}_{k}^{(t)}$.

\subsubsection{Distributional Consistency Verification}

Practical SFL systems exhibit inherently heterogeneous data, which complicates the reliable detection of malicious participants. 
To further identify malicious clients within the suspicious client set, we propose a distributional consistency verification mechanism that distinguishes adversarial behavior from benign non-IID variations.

\begin{definition}[Distributional Consistency]
	\label{def:dcs_op}
	Let ${\mathbf{Z}}(\mathcal{S}, k) = \{ z_j \mid z_j \in \mathcal{B}_{i,k}^{(t)},\, c_i \in \mathcal{S} \}$
	denote the class-$k$ features of clients in $\mathcal{S}$.
	Given two client sets $\mathcal{S}_1$ and $\mathcal{S}_2$,
	the distributional consistency of class $k$ between them is quantified as
	\begin{equation}
		\Delta(\mathcal{S}_1, \mathcal{S}_2, k) = {dist} \Big( 
		\mathcal{G}({\mathbf{Z}}(\mathcal{S}_1, k)), \,
		\mathcal{G}({\mathbf{Z}}(\mathcal{S}_2, k))
		\Big),
		\label{eq:cons_func}
	\end{equation}
\end{definition}

Based on Definition~\ref{def:dcs_op}, the impact of each suspicious client
$c_i \in \tilde{\mathcal{C}}^{(t)}_k$ on the class-wise safe zone $\mathcal{C}_k^{(t)}$
is measured by the induced distributional shift, formalized as the 
Distributional Consistency Score:
\begin{equation} 
	\text{DCS}_{i,k}^{(t)} = \Delta
	\bigl( 
	\mathcal{C}_{k}^{(t)}, \,
	\mathcal{C}_{k}^{(t)} \cup \{c_i\} \, ,k
	\bigr).
	\label{eq:dcs}
\end{equation}

To capture the intrinsic dispersion of class-$k$ features within the safe zone, we define the MP as a threshold quantifying the permissible deviations in the benign distribution:
\begin{equation} 
	\text{MP}_k^{(t)} = \max_{c_i \in \mathcal{C}_{k}^{(t)}}\Delta
	\bigl( 
	\mathcal{C}_{k}^{(t)}, \,
	\mathcal{C}_{k}^{(t)} \setminus \{c_i\}, \, k
	\bigr),
	\label{eq:mp}
\end{equation}

Suspicious clients whose scores fall within the $\text{MP}_k^{(t)}$ are regarded as benign and retained within the safe zone:
\begin{equation}
	\mathcal{C}_{k}^{(t)} \leftarrow \mathcal{C}_{k}^{(t)} \cup \{ c_i \in \tilde{\mathcal{C}}^{(t)}_{k} : \text{DCS}^{(t)}_{i,k} \le \text{MP}_k^{(t)} \}.
\end{equation}

Clients excluded from $\mathcal{C}_k^{(t)}$ are thus identified as malicious and inferred to be targeting class $k$.

\subsection{Class-Calibrated Sample Purification}

Although the first stage identifies class-wise attack targets, directly discarding all samples from these classes would severely reduce data diversity and consequently degrade model performance.
To balance robustness and data utility, we perform fine-grained purification that selectively removes poisoned samples within the identified target classes.

For each sample \( z_j \) within the identified target class \( k \), we quantify its deviation from the corresponding global class-\(k\) distribution \( \mathcal{P}_{G,k}^{(t-1)} \) as:
\begin{equation}
	D_k({z}_j) = (\phi({z_j}) - \mu_{G,k}^{(t-1)})^\top (\Sigma_{G,k}^{(t-1)})^{-1} (\phi({z_j}) - \mu_{G,k}^{(t-1)}).
	\label{eq:sample_dist}
\end{equation}

Existing methods typically rely on a single global threshold to identify anomalous samples~\cite{arp2022and,ding2025feddlad}. 
However, such thresholds fail to capture class-wise variations in feature dispersion, which can lead to misclassification. 
To address this limitation, we introduce a {class-calibrated threshold} that dynamically adjusts according to the class-wise dispersion measured by $\text{MP}_k^{(t)}$:
\begin{equation}
	\tau_k^{(t)} = Q^{(t)}_{3,k}+ \left(1 + \frac{\text{MP}^{(t)}_k - \text{MP}^{(t)}_{\min}}{\text{MP}^{(t)}_{\max} - \text{MP}^{(t)}_{\min}}\right) \, {IQR}^{(t)}_k,
	\label{eq:tau}
\end{equation}
where $\text{MP}_{\min}^{(t)} = \min_k \text{MP}_k^{(t)}$, $\text{MP}_{\max}^{(t)} = \max_k \text{MP}_k^{(t)}$, $Q^{(t)}_{3,k}$ and ${IQR}^{(t)}_k$ denote the third quartile and interquartile range  of the benign distances $\{ D_k(z_j) \mid z_j \in \mathcal{B}_{i,k}^{(t)}, c_i \in \mathcal{C}_k^{(t)} \}$.

For benign clients $c_i \in \mathcal{C}_k^{(t)}$, all class-$k$ smashed data are considered  benign: $\mathbf{Z}_{i,k}^{+} = \mathcal{B}_{i,k}^{(t)}$.
For malicious clients $c_i \notin \mathcal{C}_k^{(t)}$, the class-$k$ samples are partitioned using the threshold $\tau_k^{(t)}$ into a benign set
$\mathbf{Z}_{i,k}^{+} = \{ z_j \in \mathcal{B}_{i,k}^{(t)} \mid  D_k(z_j) \le \tau_k^{(t)} \}$
and a malicious set $\mathbf{Z}_{i,k}^{-} = \{ z_j \in \mathcal{B}_{i,k}^{(t)} \mid D_k(z_j) > \tau_k^{(t)} \}$.

Based on the validated benign set $\mathbf{Z}^{+}_{k}=\bigcup_i \mathbf{Z}_{i,k}^{+}$, we estimate a round-wise empirical distribution
$\hat{\mathcal{P}}_{k}^{(t)} = \mathcal{G}(\mathbf{Z}^{+}_{k})$, 
which is then used to update the global distribution:
\begin{equation}
	\mathcal{P}_{G,k}^{(t)} = (1-\beta)\mathcal{P}_{G,k}^{(t-1)} + \beta \hat{\mathcal{P}}_{k}^{(t)},
	\label{eq:global_update}
\end{equation}
where $\beta$ is the exponential moving average coefficient.

\subsection{Malicious Feature Decoupling}
\begin{table*}[t]
	\centering
	\caption{Performance comparison of TOFD against baselines on the MNIST dataset under Uniform and Non-IID settings ($\kappa=1$). The upper block reports accuracy (ACC $\uparrow$), while the lower block presents poisoning impact (U $\downarrow$). \textbf{Bold} and \underline{underlined} numbers indicate the best and second-best results, respectively.}
	\vspace{-2mm}
	\setlength{\tabcolsep}{4pt}
	\resizebox{\textwidth}{!}{%
		\begin{tabular}{l||cc|cc|cc|cc|cc|cc|cc}
			\toprule
			\rowcolor{lightgray}
			& \multicolumn{2}{c|}{\textbf{DP}}
			& \multicolumn{2}{c|}{\textbf{WP}}
			& \multicolumn{2}{c|}{\textbf{SP}}
			& \multicolumn{2}{c|}{\textbf{LP}}
			& \multicolumn{2}{c|}{\textbf{DP+SP}}
			& \multicolumn{2}{c|}{\textbf{WP+SP}}
			& \multicolumn{2}{c}{\textbf{LP+SP}} \\
			\rowcolor{lightgray}
			\multirow{-2}{*}{\textbf{Method}}
			& \textbf{Uniform} & \textbf{NIID}
			& \textbf{Uniform} & \textbf{NIID}
			& \textbf{Uniform} & \textbf{NIID}
			& \textbf{Uniform} & \textbf{NIID}
			& \textbf{Uniform} & \textbf{NIID}
			& \textbf{Uniform} & \textbf{NIID}
			& \textbf{Uniform} & \textbf{NIID} \\
			\specialrule{1.5pt}{0pt}{0pt}
			\multicolumn{15}{l}{Accuracy (\textbf{ACC $\uparrow$})} \\
			\midrule\midrule
			FedAvg~\cite{mcmahan2017communication}        & 10.73 & 10.29 & 18.56 & 16.21 & 91.87 & 74.95 & 60.02 & 48.53 & 10.69 & 10.66 & 65.72 & 56.27 & 43.01 & 29.34 \\
			Trim-Mean~\cite{yin2018byzantine}        & 10.28 & 10.03 & 38.31 & 32.76 & 92.59 & \underline{86.99} & 55.43 & 49.90 & 10.43 & 10.25 & 65.68 & 59.40 & 40.12 & 39.60 \\
			Median~\cite{yin2018byzantine}      & 88.85 & 85.14 & 89.34 & 85.69 & 75.13 & 64.21 & 83.00 & \underline{87.78} & 75.94 & 72.94 & 74.74 & 69.88 & 76.34 & \underline{75.22} \\
			Sparsefed~\cite{panda2022sparsefed}     & 10.65 & 10.19 & 32.44 & 14.80 & 61.33 & 42.77 & 62.54 & 53.98 & 10.89 & 10.51 & 62.20 & 52.30 & 31.93 & 20.08 \\
			Krum~\cite{blanchard2017machine}          & 91.96 & 87.15 & 91.83 & 84.03 & 77.91 & 64.61 & 89.80 & 86.44 & 78.90 & 72.90 & 84.29 & 72.91 & \underline{78.45} & 73.63 \\
			Bulyan~\cite{guerraoui2018hidden}        & 10.33 & 10.45 & 18.89 & 16.36 & 91.71 & 84.73 & 59.29 & 47.14 & 10.99 & 10.44 & 66.98 & 60.51 & 47.20 & 43.21 \\
			FLTrust~\cite{cao2020fltrust}       & {11.01} & {9.97} & 71.59 & 61.92 & 30.09 & 11.55 & 69.39 & 43.07 & 10.84 & 9.96 & 63.57 & 54.55 & 15.54 & 11.92 \\
			DnC~\cite{shejwalkar2021manipulating}          & 11.48 & 10.21 & 91.04 & 87.84 & 79.04 & 67.89 & 66.50 & 50.45 & 10.62 & 11.23 & 91.30 & \underline{86.89} & 76.51 & 67.00 \\
			ShieldFL~\cite{ma2022shieldfl}      & 11.29 & 10.78 & \underline{92.93} & \underline{90.28} & 69.10 & 55.52 & 70.54 & 67.61 & 11.36 & 11.14 & \underline{92.56} & 86.74 & 66.96 & 59.47 \\
			PRFL~\cite{yuan2025prfl}          & 89.82 & 85.19 & 90.89 & 83.09 & 78.60 & 72.88 & \underline{90.95} & 86.20 & 78.67 & 74.21 & 83.05 & 67.07 & 77.74 & 71.74 \\
			FAVD~\cite{kumar2025fortifying}          & 10.47 & 10.46 & 82.93 & 75.11 & 84.49 & 73.57 & 64.47 & 49.17 & 11.12 & 9.71 & 88.54 & 55.74 & 64.51 & 61.47 \\
			HealSplit~\cite{xie2026healsplit}      & \underline{91.28} & \underline{88.15} & 91.86 & 86.27 & \underline{93.59} & 80.86 & 67.17 & 54.56 & \underline{80.34} & \underline{75.11} & 88.19 & 81.16 & 49.53 & 36.14 \\
			\rowcolor{lightorange}
			\textbf{TOFD} & \textbf{92.92} & \textbf{90.08} & \textbf{93.95} & \textbf{91.33} & \textbf{95.35} & \textbf{92.09} & \textbf{92.69} & \textbf{89.95} & \textbf{85.81} & \textbf{81.79} & \textbf{93.35} & \textbf{88.02} & \textbf{83.70} & \textbf{80.48} \\
			\specialrule{1.5pt}{0pt}{0pt}
			\multicolumn{15}{l}{Poisoning Impact (U $\downarrow$)} \\
			\midrule\midrule
			FedAvg~\cite{mcmahan2017communication}        & 84.85 & 83.34 & 77.02 & 77.42 & 3.71 & 18.68 & 35.56 & 45.10 & 84.89 & 82.97 & 29.86 & 37.36 & 52.57 & 64.29 \\
			Trim-Mean~\cite{yin2018byzantine}        & 85.54 & 85.15 & 57.51 & 62.42 & 3.23 & 8.19 & 40.39 & 45.28 & 85.39 & 84.93 & 30.14 & 35.78 & 55.70 & 55.58 \\
			Median~\cite{yin2018byzantine}      & 3.46 & 6.38 & 2.97 & 5.83 & 17.18 & 27.31 & 9.31 & 3.74 & 16.37 & 18.58 & 17.57 & 21.64 & 15.97 & \underline{16.30} \\
			Sparsefed~\cite{panda2022sparsefed}     & 84.06 & 66.86 & 62.27 & 62.25 & 33.38 & 34.28 & 32.17 & 23.07 & 83.82 & 66.54 & 32.51 & 24.75 & 62.78 & 56.97 \\
			Krum~\cite{blanchard2017machine}          & \textbf{1.96} & \textbf{1.06} & \underline{2.09} & 4.18 & 16.01 & 23.60 & 4.12 & \underline{1.77} & \underline{15.02} & 15.31 & 9.63 & 15.30 & \underline{15.47} & 14.58 \\
			Bulyan~\cite{guerraoui2018hidden}        & 85.31 & 83.16 & 76.75 & 77.25 & 3.93 & 8.88 & 36.35 & 46.47 & 84.65 & 83.17 & 28.66 & 33.10 & 48.44 & 50.40 \\
			FLTrust~\cite{cao2020fltrust}       & 84.93 & 83.55 & 24.35 & 31.60 & 65.85 & 81.97 & 26.55 & 50.45 & 85.10 & 83.56 & 32.37 & 38.97 & 80.40 & 81.60 \\
			DnC~\cite{shejwalkar2021manipulating}          & 83.43 & 82.04 & 3.87 & 4.41 & 15.87 & 24.36 & 28.41 & 41.80 & 84.29 & 81.02 & 3.61 & \underline{5.36} & 18.40 & 25.25 \\
			ShieldFL~\cite{ma2022shieldfl}      & 83.99 & 81.66 & 2.35 & \underline{2.16} & 26.18 & 36.92 & 24.74 & 24.83 & 83.92 & 81.30 & \underline{2.72} & 5.70 & 28.32 & 32.97 \\
			PRFL~\cite{yuan2025prfl}          & 4.06 & 3.66 & 2.99 & 5.76 & 15.28 & 15.97 & \textbf{2.93} & 2.65 & 15.21 & \underline{14.64} & 10.83 & 21.78 & 16.14 & 17.11 \\
			FAVD~\cite{kumar2025fortifying}          & 84.42 & 78.22 & 11.96 & 13.57 & 10.40 & 15.11 & 30.42 & 39.51 & 83.77 & 78.97 & 6.35 & 32.94 & 30.38 & 27.21 \\
			HealSplit~\cite{xie2026healsplit}      & 4.22 & 5.13 & 3.64 & 7.01 & \underline{1.91} & 12.42 & 28.33 & 38.72 & 15.16 & 18.17 & 7.31 & 12.12 & 45.97 & 57.14 \\
			\rowcolor{lightorange}
			\textbf{TOFD} & \underline{3.10} & \underline{2.40} & \textbf{2.07} & \textbf{1.15} & \textbf{0.67} & \textbf{0.39} & \underline{3.33} & \textbf{2.53} & \textbf{10.21} & \textbf{10.69} & \textbf{2.67} & \textbf{4.46} & \textbf{12.32} & \textbf{12.00} \\
			\specialrule{1.5pt}{0pt}{0pt}
	\end{tabular}}
	\label{tab:comparison_combined}
\end{table*}

Despite effective filtering of smashed data, client-side models trained under adversarial control can still introduce malicious patterns into the global model through aggregation. 
To mitigate these residual effects, we propose a decoupling mechanism for adversarial features.

%Despite effective filtering of smashed data, client-side models trained under adversarial control can still introduce malicious patterns into the global model through aggregation.

Specifically, an adversarial guidance model $\mathcal{R}$, with the same architecture as the server-side model $h_{\theta_s}$, is fine-tuned on the detected malicious smashed data \(\mathbf{Z}^- = \bigcup_{i,k} \mathbf{Z}^-_{i,k}\) in each communication round to capture attack-induced patterns. 
After convergence, $\mathcal{R}$ serves as a stable module across rounds. 
To enforce decoupling, the server-side model \(h_{\theta_{s_i}}\) is encouraged to diverge from \(\mathcal{R}\) by maximizing the predictive discrepancy on \(\mathbf{Z}^{-}\), yielding the following objective:
\begin{equation}
	\mathcal{L}_{\text{MFD}}
	= - \frac{1}{|{\mathbf{Z}}^-|}
	\sum_{z \in {\mathbf{Z}}^-}
	\mathrm{KL}\left(
	f(z;\mathcal{R}) \;\middle\|\; f(z; h_{\theta_{s_i}})
	\right),
\end{equation}
where $\mathrm{KL}$ denotes the Kullback–Leibler divergence, and $f(\cdot; \cdot)$ represents the softmax output of the model.

The server-side model of client $c_i$ is trained on the validated benign smashed data using the  cross-entropy loss:
\begin{equation}
	\mathcal{L}_{\text{CE}}
	= \frac{1}{|\bigcup_k \mathbf{Z}_{i,k}^+|} \sum_{k=1}^K 
	\sum_{z \in  \mathbf{Z}_{i,k}^+} - y_{(k)}^\top \log f(z; h_{\theta_{s_i}}).
\end{equation}
where $y_{(k)}$ is the one-hot label vector for class $k$.

%\paragraph{Joint Loss Function.} 
The overall optimization objective for each participating client combines task supervision with adversarial decoupling:
\begin{equation}
	\mathcal{L}_{\text{total}}
	= \mathcal{L}_{\text{CE}} + \lambda \, \mathcal{L}_{\text{MFD}},
\end{equation}
where $\lambda$ balances task performance with suppression of residual adversarial influence.

\section{Theoretical Analysis of TOFD}
\label{sec:theory}

In this section, we present the analysis of the time complexity and convergence guarantees of TOFD, with detailed proofs deferred to {Appendix~A}.

\begin{assumption}
\label{ass:rep_lip}
\textbf{Lipschitz Smoothness.}
The gradient $\nabla_\theta \ell(\theta, z)$ is $L_z$-Lipschitz with respect to the smashed data $z$, i.e, 
$$\|\nabla_\theta \ell(\theta, z_1) - \nabla_\theta \ell(\theta, z_2)\| \le L_z \|z_1 - z_2\|$$
\end{assumption}

\begin{algorithm}[t]
	\caption{Framework of TOFD}
	\label{alg:tofd}
	\begin{algorithmic}[1]
		\STATE {\bfseries Input:} current round \(t\), number of classes \(K\), Clients \(\mathcal{C}\), smashed data $\{\mathcal{B}_{i,k}^{(t)}\}_{i=1,j=1}^{N, K}$, and parameters \(\lambda, \beta\)
		%$\{\mathcal{B}_{i,k}^{(t)}\}_{i\in [N],k \in [K]}$
		\FOR{each class $k \in \{1, \dots, K\}$}
		\STATE{Initialize malicious set $\mathbf{Z}_{i,k}^{-}\leftarrow \emptyset$, benign set $\mathbf{Z}_{i,k}^{+}\leftarrow \emptyset$} for each participating clients 
		\FOR{each client $c_i \in \mathcal{C}$}
		\STATE Calculate distribution distance $\mathcal{W}_{i,k}^{(t)}$ using Eq.\eqref{eq:dist_gaussians}   
		\ENDFOR
		\STATE Construct initial safe zone $\mathcal{C}_{k}^{(t)}$ based on $\{\mathcal{W}_{i,k}^{(t)}\}_{c_i \in \mathcal{C}}$
		\STATE Compute $\text{MP}_k^{(t)}$ using $\mathcal{C}_{k}^{(t)}$ via Eq.\eqref{eq:mp}
		
		\FOR{each suspicious client $c_i \notin \mathcal{C}_{k}^{(t)}$}
		%     \STATE{Calculate $\text{DCS}_{i,k}^{(t)}$ via Eq. \eqref{eq:dcs}}
		\STATE \textbf{if} $\text{DCS}_{i,k}^{(t)} \le \text{MP}_k^{(t)}$ \textbf{then} add $c_i$ to $\mathcal{C}_{k}^{(t)}$
		\ENDFOR
		
		% \FOR{each sample $z_j$ in $\bigcup_{c_i \notin \mathcal{C}_{k}^{(t)}} \mathcal{B}_{i,k}^{(t)}$}
		% \FOR{each sample $z_j \in \mathcal{B}_{i,k}^{(t)}$ s.t. $c_i \notin \mathcal{C}_{k}^{(t)}$}
		% \FOR{each sample  $\{ z_j \in \mathcal{B}_{i,k}^{(t)} \mid c_i \notin \mathcal{C}_{k}^{(t)} \}$}
		\FOR{each $z_j \in \mathcal{B}_{i,k}^{(t)}$ where $c_i \notin \mathcal{C}_{k}^{(t)}$}
		\STATE \textbf{if} $D_k(z_j)>\tau_k^{(t)}$ \textbf{then} add $z_j$ to $\mathbf{Z}_{i,k}^{-}$
		%$\mathbf Z_k^{-}\leftarrow \mathbf Z_k^{-}\cup\{z_j\}$
		\ENDFOR
		\STATE Set $\mathbf Z_{i,k}^{+}\leftarrow
		\mathcal B_{i,k}^{(t)}\setminus \mathbf Z_{i,k}^{-}$
		\STATE Update $\mathcal{P}_{G,k}^{(t)}$ using $\mathbf{Z}_k^{+}= \bigcup_{i} \mathbf{Z}_{i,k}^{+}$ via Eq.\eqref{eq:global_update} 
		\ENDFOR
		\STATE Fine-tune model $\mathcal{R}$ on $\mathbf{Z}^{-}=\bigcup_{i,k} \mathbf{Z}_{i,k}^{-}$
		\FOR{client $c_i$ in $\mathcal{C}$}
		\STATE Optimize SFL on $\mathbf{Z}_i^{+} =\bigcup_k \mathbf{Z}_{i,k}^{+}$ using $\mathcal{R}$ and $\mathcal{L}_{\text{total}}$ 
		\ENDFOR
		\STATE \textbf{return} Aggregated global model $f_{\theta}$
	\end{algorithmic}
\end{algorithm}

\begin{assumption}
\label{ass:reconstruct}
\textbf{Reconstruction Error.}
Let $\psi(\cdot)$ denote an operator that maps compressed smashed data back to the original feature space. For benign smashed data $z^+$, the reconstruction error satisfies $|z^+-\psi(\phi(z^+))|\le\sigma_{\mathrm{sp}}$.
\end{assumption}

\begin{assumption}
\label{ass:ref_stability}
\textbf{Representation Covariance.}
For each class $k$, let $\Sigma_{k}^{(t)}$ denote the covariance matrix of compressed benign smashed data at round $t$. We assume
$$\sup_{t} \lambda_{\max}\!\left(\Sigma_{k}^{(t)}\right) \le \Lambda_k < \infty.$$
\end{assumption}

\begin{assumption}\textbf{Discard-Induced Deviation.}
\label{ass:trunc_decay}
Let $\delta$ denote the mean deviation induced by discarding benign samples whose distance exceeds $\tau_k$. We assume that
$$\lim_{\tau_k \to \infty} \delta = 0.$$

\end{assumption}

\begin{lemma}[Time-Complexity Analysis]
\label{lem:time_complexity}
The time complexity of each server-side training step is 
$\mathcal{O}(|\mathcal{D}^\mathrm{tr}_i| \cdot T \cdot N_{\mathrm{p}})$, 
where $N_{\mathrm{p}}$ is the number of model parameters. 
TOFD introduces an additional cost of 
$\mathcal{O}(|\mathbf{Z}^{-}| \cdot T_{\mathcal{R}} \cdot N_{\mathrm{p}})$, 
where $T_{\mathcal{R}}$ denotes the number of training rounds for $\mathcal{R}$. 
In typical SFL settings, the additional overhead is negligible because 
$|\mathbf{Z}^{-}| \ll |\mathcal{D}_i^\mathrm{tr}|$ and $T_{\mathcal{R}} \ll T$.
\end{lemma}

\begin{lemma}[Bounded Gradient Bias]
\label{lem:bias_main}
Let $\alpha \in [0,1)$ denote an upper bound on the fraction of undetected malicious samples, 
and let $S$ denote the compressed dimension of the smashed data after applying $\phi(\cdot)$ during detection. Under assumptions \ref{ass:rep_lip} to \ref{ass:trunc_decay}, the gradient bias satisfies:
\begin{equation*}
\label{eq:bias_bound}
\|b_t\| \le \varepsilon := \max_k \sqrt{S} L_z \left(  \sqrt{S \Lambda_k} +\delta
+ \frac{1}{1-\alpha}\,\sqrt{\tau_k \Lambda_k} \right)+ 4L_z\sigma_{sp} ,
\end{equation*}
% where $L_z$ denotes the Lipschitz constant, $\sigma_{\mathrm{sp}}$ is the reconstruction error, $\delta$ is the truncation bias, $\tau_k$ is the maximum $\tau_k^{t}$ over all $t$, and $\Lambda_k$ is the reference covariance bound.
\end{lemma}

Building on Lemma~\ref{lem:bias_main} and assumptions, 
the following convergence guarantee is established for TOFD.

\begin{theorem}[Convergence of TOFD]
\label{theo:convergence}
With stepsize $\eta=\mathcal{O}(1/\sqrt{T})$, the iterates  $\{\theta_t\}_{t=0}^{T-1}$ produced by TOFD satisfy
\[\frac{1}{T}\sum_{t=0}^{T-1}\mathbb{E}\|\nabla Q(\theta_t)\|^2
\le
\mathcal{O}\!\left(\frac{1}{\sqrt{T}}\right)+\mathcal{O}\!\left(\varepsilon^2\right).
\]
\end{theorem}

\section{Evaluation}\label{Evaluation}

\begin{figure}[t]
    \centering   \includegraphics[width=0.95\linewidth]{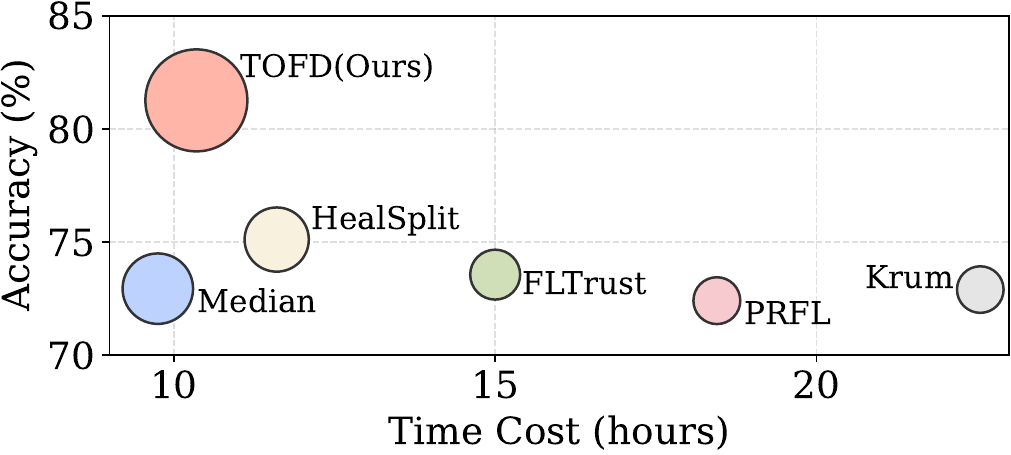}
        \vspace{-2mm}
    \caption{Comparison of accuracy and time cost.}
    \label{fig:time_acc}
\end{figure}
\begin{figure*}[t]
	\centering	\includegraphics[width=1\textwidth]{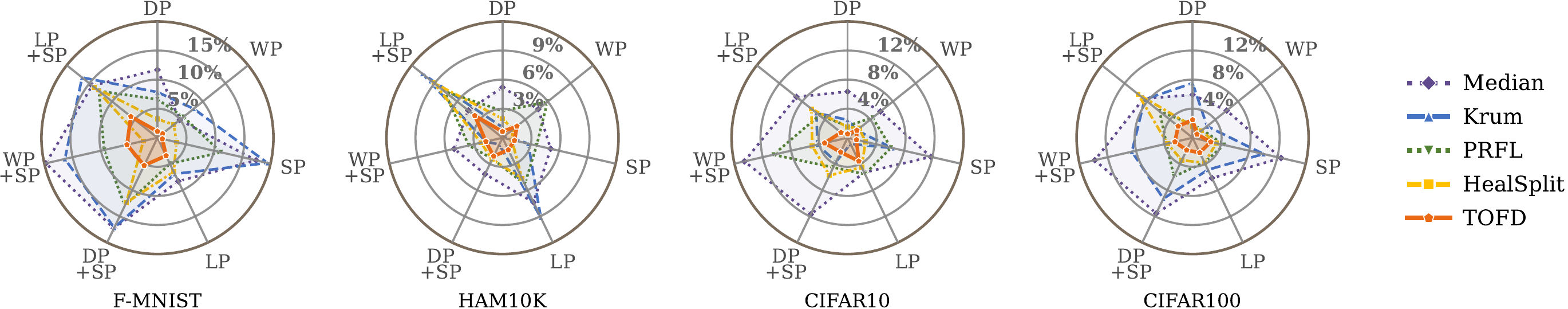}
	    \vspace{-6mm}
    \caption{Comparison of metric $U \downarrow$ across datasets for different defense strategies under various attacks.}
	\label{fig:radar}
\end{figure*}

\begin{figure*}[t]
\centering
\begin{minipage}{0.3\textwidth}
	\centering
	\includegraphics[width=\linewidth]{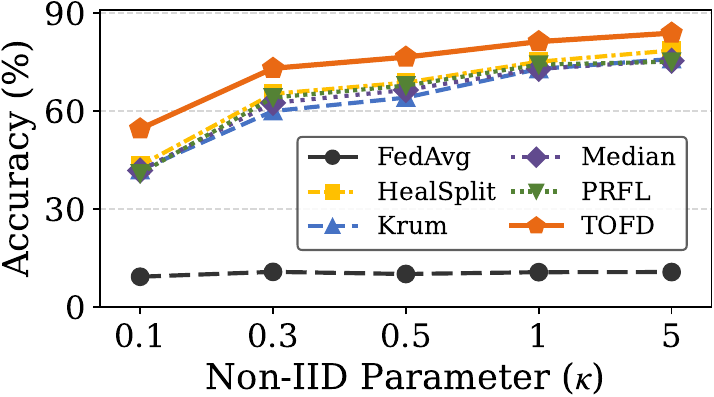}
    \vspace{-6.8mm}
	\caption{Performance under different Non-IID degrees ($\kappa$).}
	\label{fig:niid}
\end{minipage}
\hfill\hfill
\begin{minipage}{0.31\textwidth}
	\centering	\includegraphics[width=\linewidth]{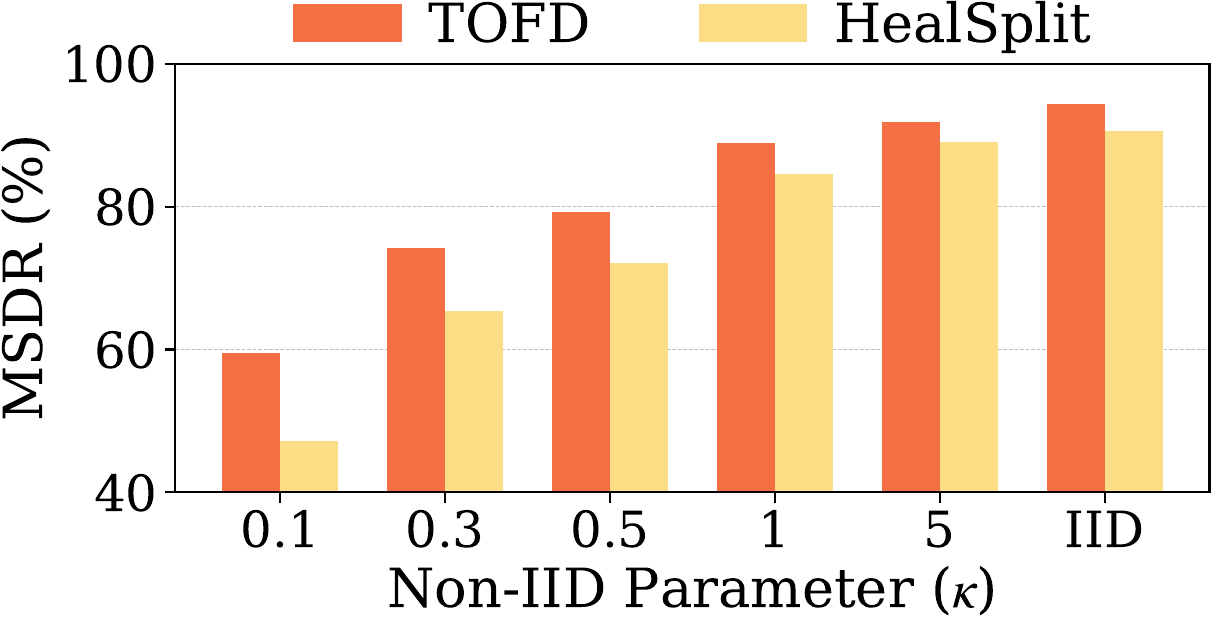}
        \vspace{-6mm}
	\caption{Comparison of MSDR under different Non-IID degrees ($\kappa$).}
	\label{fig:msdr}
\end{minipage}
\hfill\hfill
\begin{minipage}{0.31\textwidth}
	\centering
\includegraphics[width=\linewidth]{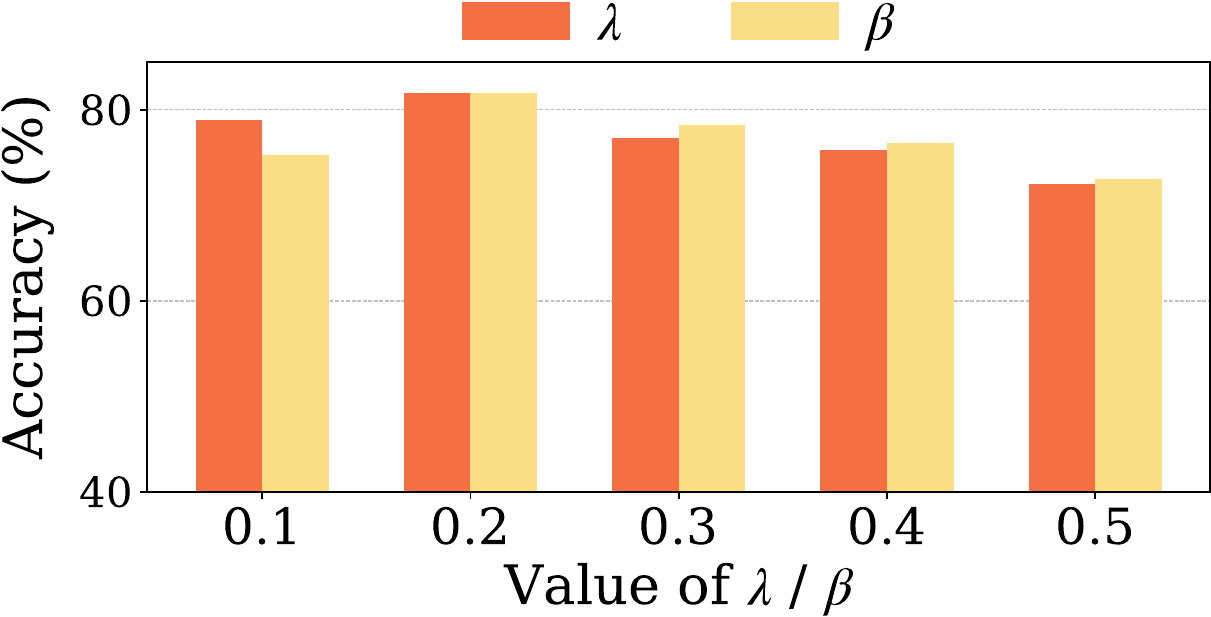}
     \vspace{-6mm}
    \caption{Performance under different hyperparameters ($\beta$ and $\lambda$).}
	\label{fig:acc_hyper}
\end{minipage}
\end{figure*}

\subsection{Experiment Setup}\label{Experiment_Setup}

\paragraph{Datasets and Baselines: }
TOFD is evaluated on five image classification benchmarks: MNIST~\cite{lecun1998mnist}, Fashion-MNIST~\cite{xiao2017fashion}, HAM10k~\cite{tschandl2018ham10000}, CIFAR10~\cite{krizhevsky2009learning}, and CIFAR-100. To simulate realistic SFL scenarios, client data are partitioned in a non-IID manner according to a Dirichlet distribution $Dir(\kappa)$, with smaller values of $\kappa$ corresponding to higher data heterogeneity. Experiments are performed using three widely adopted backbone architectures: DenseNet121~\cite{huang2017densely}, ResNet-18~\cite{he2016deep}, and ResNet-50.

To assess TOFD's robustness, we evaluate individual attack strategies including DP, WP, SP, and LP, as well as more complex combinations such as DP + SP, WP + SP, and LP + SP. TOFD is then compared with a diverse set of state-of-the-art defense methods, including FedAvg~\cite{mcmahan2017communication}, Trimmed-Mean, Median, SparseFed~\cite{panda2022sparsefed}, Krum ~\cite{blanchard2017machine}, Bulyan~\cite{guerraoui2018hidden}, FLTrust~\cite{cao2020fltrust}, DnC~\cite{shejwalkar2021manipulating}, ShieldFL~\cite{ma2022shieldfl}, PRFL~\cite{yuan2025prfl}, FAVD~\cite{kumar2025fortifying}, and HealSplit~\cite{xie2026healsplit} .

\paragraph{Settings and Metrics}
To simulate realistic SFL systems, we vary the experimental settings by adjusting key factors such as dataset, model architecture, and data heterogeneity. A summary of the default configurations used in our experiments is provided in Table~\ref{tab:setting}. All experiments were conducted on an NVIDIA A100 GPU.

The experimental results are evaluated using four metrics: 
{test accuracy} $\mathcal{A}$, which reflects the defense performance; 
{the poisoning impact} $U = \mathcal{A}^* - \mathcal{A}$, $\mathcal{A}^*$ measuring the performance degradation caused by poisoning attacks in the presence of the defense, where $\mathcal{A}^*$ denotes the accuracy without attacks; 
{MSDR}, which measures the rate of detected malicious smashed data; and time cost, which quantifies the computational overhead of the defense mechanism.

\subsection{Evaluation Results}

\subsubsection{Robustness under Diverse Attacks.}
Our first set of experiments evaluates TOFD under various poisoning attacks in both IID and non-IID settings. Table~\ref{tab:comparison_combined} reports the test accuracy and poisoning impact under different attack scenarios. Figure~\ref{fig:time_acc} further presents the computational cost of state-of-the-art methods.

TOFD consistently achieves strong defense performance across diverse attack types and remains robust under composite attacks, maintaining over 92\% accuracy under single attacks and 83\% under composite attacks in the uniform setting. Unlike most coarse-grained defenses that discard entire client updates, TOFD filters poisoned smashed data at a fine-grained level while preserving informative benign samples. This advantage is particularly evident under composite attacks. For example, Trimmed-Mean only achieves 10.43\% accuracy under DP+SP attacks in uniform settings, with a high poisoning impact of 85.39\%, whereas TOFD still maintains 85.81\% accuracy with a much lower poisoning impact of 10.21\%, demonstrating its robustness under complex attack scenarios.

TOFD also maintains strong defense performance under heterogeneous data settings. Under the WP+SP attack, shifting from IID to non-IID causes FAVD’s accuracy to drop sharply, with the poisoning impact rising by 26.59\%. In contrast, TOFD maintains high accuracy, while the poisoning impact increases by only 1.79\%. This improvement arises because baseline methods relying on fixed criteria often fail to distinguish benign non-IID variations from malicious samples, whereas TOFD employs adaptive, class-specific metrics to effectively isolate adversarial behavior. Meanwhile, TOFD maintains competitive computational cost while achieving the highest average accuracy among all methods, indicating a favorable trade-off between defense effectiveness and runtime efficiency for practical SFL systems.

\subsubsection{Defense Generalization Across Data.} 
Our second set of experiments investigates TOFD’s performance under different attack strategies across four datasets. Figure~\ref{fig:radar} summarizes the poisoning impact across these datasets, highlighting the consistency of performance under diverse data distributions.

We analyze the generalization of representative defenses across four datasets with default heterogeneous class distributions. TOFD achieves superior performance across all datasets compared to the state-of-the-art baseline. While Median and Krum show moderate resilience, their fixed thresholds struggle to accommodate dataset-specific variations, and PRFL and HealSplit, despite incorporating sample purification, still exhibit residual vulnerability under complex attacks. In contrast, TOFD adaptively calibrates its detection thresholds based on observed class-wise statistics, allowing it to account for differences in class distributions across datasets. This class-adaptive mechanism effectively separates malicious perturbations from benign variations, enabling TOFD to maintain consistently low deviations and robust defense performance across diverse datasets in heterogeneous SFL settings.

\subsubsection{Robustness under Varying Data Heterogeneity.}
Our third set of experiments evaluates the robustness of TOFD under varying degrees of data heterogeneity. The test accuracy and MSDR results are reported in Figures~\ref{fig:niid} and~\ref{fig:msdr}, respectively.

As non-IID heterogeneity increases, TOFD’s performance advantage becomes more pronounced. While baseline defenses degrade significantly under stronger heterogeneity, TOFD consistently maintains stable accuracy across all levels.
This trend is further supported by the MSDR results. TOFD consistently achieves higher MSDR than HealSplit across all heterogeneity levels, and the performance gap widens under stronger heterogeneity, demonstrating its ability to handle heterogeneous data distributions effectively.

\subsubsection{Hyperparameters and Attack Intensity} 
Our fourth set of experiments assesses TOFD’s sensitivity to hyperparameter choices and malicious client ratio.
Results are shown in Figure~\ref{fig:acc_hyper} and Table~\ref{tab:ratio}.

\begin{table}[t]
    \centering
    \caption{Accuracy (\%) of different methods under different malicious client ratio.}
    \label{tab:ratio}
        \vspace{-2mm}
    \resizebox{0.85\columnwidth}{!}{%
    \begin{tabular}{lccccc}
        \toprule
        \rowcolor{lightgray}  
        & \multicolumn{5}{c}{\textbf{Malicious Client Ratio}} \\
        \rowcolor{lightgray}
        \multirow{-2}{*}{\textbf{Method}}
        & \textbf{5\%} & \textbf{10\%} & \textbf{15\%} & \textbf{20\%} & \textbf{25\%} \\
        \midrule
        Median~\cite{yin2018byzantine}    & 87.85 & 82.37 & 77.24 & 72.94 & 71.19 \\
        Krum \cite{blanchard2017machine}     & 85.09 & 82.54 & 77.81 & 72.90 & 70.02 \\
        PRFL \cite{yuan2025prfl}      & 86.69 & 83.09 & 78.47 & 74.21 & 72.57 \\
        HealSplit \cite{xie2026healsplit}  & 89.79 & 86.08 & 81.00 & 75.11 & 73.79 \\
        \rowcolor{lightorange}
        \textbf{TOFD} & \textbf{91.04} & \textbf{88.50} & \textbf{85.38} & \textbf{81.79} & \textbf{79.09} \\
        \bottomrule
    \end{tabular}%
    }
          \vspace{-2mm}
\end{table}

We first investigate the impact of the decoupling weight $\lambda$ and the EMA coefficient $\beta$ on performance. TOFD demonstrates stable test accuracy over a broad range of values, indicating robustness to hyperparameter variations. Varying $\lambda$ results in peak performance at $\lambda = 0.2$, with neighboring values yielding comparable results. A similar pattern is observed for $\beta$. Based on these findings, we adopt $\lambda = \beta = 0.2$ as the default configuration for all experiments.

To further evaluate TOFD’s robustness under practical attack scenarios, we investigate its performance across different fractions of malicious clients. As the ratio of malicious clients increases from 5\% to 25\%, TOFD maintains the highest accuracy with minimal performance drop, unlike baseline methods that suffer significant degradation, demonstrating strong resilience to intensified adversarial participation.

\subsubsection{Ablation Study.}
Our fifth experiment conducts an ablation study on both uniform and non-IID distributions to quantify the contribution of each TOFD component. The accuracy results are summarized in Table \ref{tab:ablation}.

With the complementary contributions of its components, TOFD exhibits amplified benefits under heterogeneous data distributions. Among all components, removing the purification module, which discards entire malicious clients, results in the largest performance drop, especially under non-IID distributions, underscoring the importance of fine-grained filtering. Ablating $\text{MP}_k$ or $\tau_k$ individually also degrades accuracy, as replacing adaptive, class-specific thresholds with global averages eliminates class-wise adaptation and weakens the separation between adversarial and benign variations. Removing the malicious feature decoupling loss $\mathcal{L}_{\text{MFD}}$ leads to a smaller yet consistent decline, confirming its role in mitigating residual adversarial effects.

\subsubsection{Adaptive Attack.}

\begin{figure}[t]
    \centering       \includegraphics[width=0.9\linewidth]{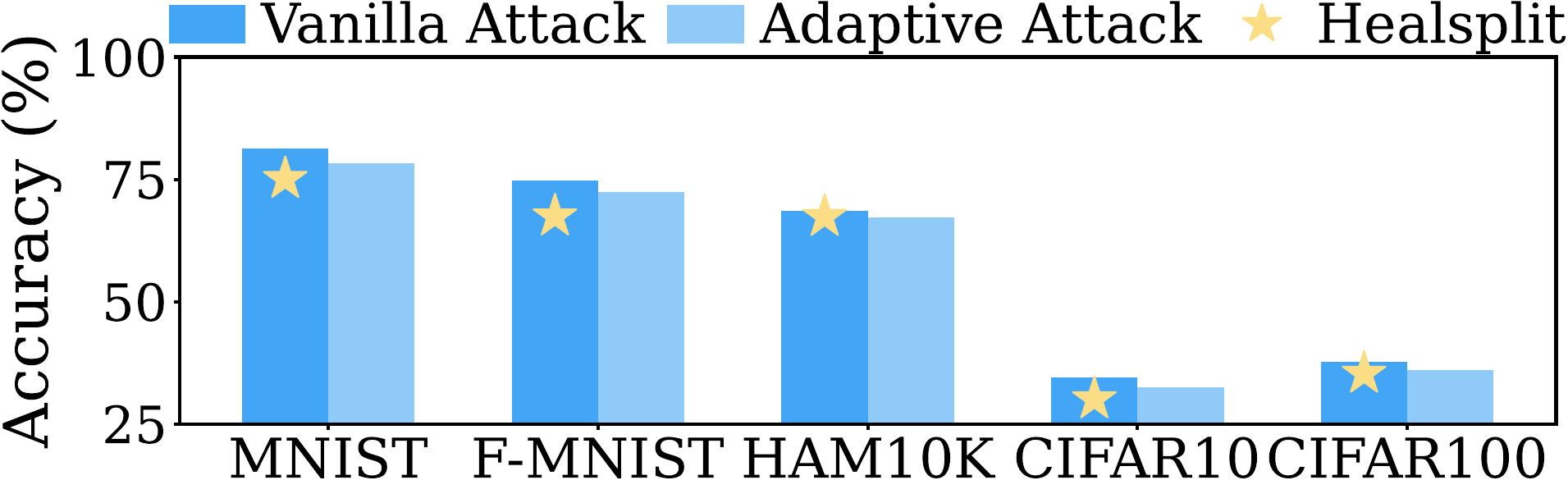}
    \vspace{-2mm}
    \caption{Comparison of adaptive attacks.}

    \label{fig:adaptive}
\end{figure}

The sixth set of experiments evaluates the resilience of TOFD against adaptive attacks across five datasets. The accuracy results are presented in Fig. \ref{fig:adaptive}.

In this scenario, the attacker constrains the induced DCS within the margin perturbation $\text{MP}_k$, exploiting benign non-IID variability to bypass distributional consistency verification. These subtle deviations partially evade threshold-based sample purification and weaken malicious feature decoupling, leaving residual adversarial patterns in the SFL system that accumulate through aggregation, gradually impacting overall performance. Despite this intensified threat, TOFD maintains a clear advantage, consistently outperforming the strongest baselines. This robustness stems from TOFD’s class-adaptive mechanism, which dynamically calibrates defense based on class-wise statistics rather than a fixed global criterion, enabling effective separation of malicious perturbations from benign non-IID variations and suppressing their propagation through 
aggregation.

\begin{table}[t]
    \centering
    \footnotesize
    \caption{Results of ablation study.}
         \vspace{-2mm}
    \label{tab:ablation}
        \resizebox{\columnwidth}{!}{%
    \begin{tabular}{lcccc}
        \toprule
        \rowcolor{lightgray}
        % \multirow{2}{*}[-0.5ex]{\textbf{Component}}   
        & \multicolumn{2}{c}{\textbf{MNIST}} 
        & \multicolumn{2}{c}{\textbf{F-MNIST}} \\
        \rowcolor{lightgray}
        % \multicolumn{1}{l|}{}
        \multirow{-2}{*}{\textbf{Component}} 
        & \textbf{Uniform} & \textbf{NIID} 
        & \textbf{Uniform} & \textbf{NIID} \\
        \midrule
        \rowcolor{lightorange}
        \textbf{TOFD} & \textbf{85.81$\pm$0.8} & \textbf{81.79$\pm$0.6}
        & \textbf{77.97$\pm$0.8} & \textbf{74.83$\pm$0.9} \\
        w/o Purification & 77.20$\pm$0.7 & 71.53$\pm$1.1
        & 68.71$\pm$0.8 & 63.55$\pm$1.3 \\
        w/o $\text{MP}_k$ & 80.50$\pm$0.3 & 74.48$\pm$0.5
        & 73.86$\pm$0.9 & 69.23$\pm$0.7 \\
        w/o $\tau_k$ & 83.70$\pm$0.6 & 78.54$\pm$0.8
        & 75.15$\pm$0.4 & 71.60$\pm$0.5 \\
        w/o $\mathcal{L}_{\text{MFD}}$ & 84.39$\pm$0.9 & 79.98$\pm$0.7
        & 76.39$\pm$1.0 & 72.83$\pm$0.7 \\
        \bottomrule
    \end{tabular}}
\end{table}

\section{Conclusion} \label{conclu}
In this work, we propose {TOFD}, a unified defense framework against diverse poisoning attacks for SFL. Existing defenses largely adapt conventional FL strategies and thus fail to leverage the split architecture for early-stage intervention. TOFD addresses this limitation by integrating detection and optimization to protect SFL through securing smashed data. It first performs class-wise target inference to identify attacked classes by constructing refined safe zones, and then selectively filters malicious smashed data within these classes. Building on this detection stage, TOFD mitigates residual adversarial influence by incorporating an adversarial guidance model to enforce a decoupling objective during SFL optimization. Extensive experiments show that TOFD consistently outperforms state-of-the-art defenses in robustness and efficiency across diverse attacks, offering a practical and generalizable solution for secure SFL.
\section*{Acknowledgment}
This work was supported by the National Key R\&D Program of China (2023YFA1009500).

\bibliographystyle{ACM-Reference-Format}
\balance
\bibliography{sample-base}

@inproceedings{ding2025feddlad,
  title={FedDLAD: A Federated Learning Dual-Layer Anomaly Detection Framework for Enhancing Resilience Against Backdoor Attacks},
  author={Ding, Binbin and Yang, Penghui and Huang, Sheng-Jun},
  booktitle={Proceedings of the Thirty-Fourth International Joint Conference on Artificial Intelligence, IJCAI-25},
  pages={5021--5029},
  year={2025}
}

@inproceedings{arp2022and,
  title={Dos and don'ts of machine learning in computer security},
  author={Arp, Daniel and Quiring, Erwin and Pendlebury, Feargus and Warnecke, Alexander and Pierazzi, Fabio and Wressnegger, Christian and Cavallaro, Lorenzo and Rieck, Konrad},
  booktitle={31st USENIX Security Symposium (USENIX Security 22)},
  pages={3971--3988},
  year={2022}
}

@inproceedings{shejwalkar2021manipulating,
  title={Manipulating the byzantine: Optimizing model poisoning attacks and defenses for federated learning},
  author={Shejwalkar, Virat and Houmansadr, Amir},
  booktitle={NDSS},
  year={2021}
}

@article{ma2022shieldfl,
  title={ShieldFL: Mitigating model poisoning attacks in privacy-preserving federated learning},
  author={Ma, Zhuoran and Ma, Jianfeng and Miao, Yinbin and Li, Yingjiu and Deng, Robert H},
  journal={IEEE Transactions on Information Forensics and Security},
  volume={17},
  pages={1639--1654},
  year={2022},
  publisher={IEEE}
}

@article{li2024introducing,
  title={Introducing edge intelligence to smart meters via federated split learning},
  author={Li, Yehui and Qin, Dalin and Poor, H Vincent and Wang, Yi},
  journal={Nature communications},
  volume={15},
  number={1},
  pages={9044},
  year={2024},
  publisher={Nature Publishing Group UK London}
}

@inproceedings{li2025infighting,
  title={Infighting in the Dark: Multi-Label Backdoor Attack in Federated Learning},
  author={Li, Ye and Zhao, Yanchao and Zhu, Chengcheng and Zhang, Jiale},
  booktitle={Proceedings of the Computer Vision and Pattern Recognition Conference},
  pages={25770--25779},
  year={2025}
}

@inproceedings{fang2020local,
  title={Local model poisoning attacks to Byzantine-Robust federated learning},
  author={Fang, Minghong and Cao, Xiaoyu and Jia, Jinyuan and Gong, Neil},
  booktitle={29th USENIX security symposium (USENIX Security 20)},
  pages={1605--1622},
  year={2020}
}

@inproceedings{sandeepa2024sherpa,
  title={Sherpa: Explainable robust algorithms for privacy-preserved federated learning in future networks to defend against data poisoning attacks},
  author={Sandeepa, Chamara and Siniarski, Bartlomiej and Wang, Shen and Liyanage, Madhusanka},
  booktitle={2024 IEEE Symposium on Security and Privacy (SP)},
  pages={4772--4790},
  year={2024},
  organization={IEEE}
}

@article{jha2023label,
  title={Label poisoning is all you need},
  author={Jha, Rishi and Hayase, Jonathan and Oh, Sewoong},
  journal={Advances in Neural Information Processing Systems},
  volume={36},
  pages={71029--71052},
  year={2023}
}

@inproceedings{jiangfedclean,
  title={FedClean: A General Robust Label Noise Correction for Federated Learning},
  author={Jiang, Xiaoqian and Zhang, Jing},
  year={2025},
  booktitle={Forty-second International Conference on Machine Learning}
}

@article{yazdinejad2024robust,
  title={A robust privacy-preserving federated learning model against model poisoning attacks},
  author={Yazdinejad, Abbas and Dehghantanha, Ali and Karimipour, Hadis and Srivastava, Gautam and Parizi, Reza M},
  journal={IEEE Transactions on Information Forensics and Security},
  volume={19},
  pages={6693--6708},
  year={2024},
  publisher={IEEE}
}

@inproceedings{mcmahan2017communication,
  title={Communication-efficient learning of deep networks from decentralized data},
  author={McMahan, Brendan and Moore, Eider and Ramage, Daniel and Hampson, Seth and y Arcas, Blaise Aguera},
  booktitle={Artificial intelligence and statistics},
  pages={1273--1282},
  year={2017},
  organization={PMLR}
}

@inproceedings{panda2022sparsefed,
  title={Sparsefed: Mitigating model poisoning attacks in federated learning with sparsification},
  author={Panda, Ashwinee and Mahloujifar, Saeed and Bhagoji, Arjun Nitin and Chakraborty, Supriyo and Mittal, Prateek},
  booktitle={International Conference on Artificial Intelligence and Statistics},
  pages={7587--7624},
  year={2022},
  organization={PMLR}
}

@article{lin2024efficient,
  title={Efficient parallel split learning over resource-constrained wireless edge networks},
  author={Lin, Zheng and Zhu, Guangyu and Deng, Yiqin and Chen, Xianhao and Gao, Yue and Huang, Kaibin and Fang, Yuguang},
  journal={IEEE Transactions on Mobile Computing},
  volume={23},
  number={10},
  pages={9224--9239},
  year={2024},
  publisher={IEEE}
}

@article{xiao2017fashion,
  title={Fashion-mnist: a novel image dataset for benchmarking machine learning algorithms},
  author={Xiao, Han and Rasul, Kashif and Vollgraf, Roland},
  journal={arXiv preprint arXiv:1708.07747},
  year={2017}
}

@article{krizhevsky2009learning,
  title={Learning multiple layers of features from tiny images},
  author={Krizhevsky, Alex and Hinton, Geoffrey and others},
  year={2009},
  publisher={Toronto, ON, Canada}
}

@article{tschandl2018ham10000,
  title={The HAM10000 dataset, a large collection of multi-source dermatoscopic images of common pigmented skin lesions},
  author={Tschandl, Philipp and Rosendahl, Cliff and Kittler, Harald},
  journal={Scientific data},
  volume={5},
  number={1},
  pages={1--9},
  year={2018},
  publisher={Nature Publishing Group}
}

@inproceedings{huang2017densely,
  title={Densely connected convolutional networks},
  author={Huang, Gao and Liu, Zhuang and Van Der Maaten, Laurens and Weinberger, Kilian Q},
  booktitle={Proceedings of the IEEE conference on computer vision and pattern recognition},
  pages={4700--4708},
  year={2017}
}

@inproceedings{he2016deep,
  title={Deep residual learning for image recognition},
  author={He, Kaiming and Zhang, Xiangyu and Ren, Shaoqing and Sun, Jian},
  booktitle={Proceedings of the IEEE conference on computer vision and pattern recognition},
  pages={770--778},
  year={2016}
}

@article{wu2024evaluating,
  title={Evaluating security and robustness for split federated learning against poisoning attacks},
  author={Wu, Xiaodong and Yuan, Henry and Li, Xiangman and Ni, Jianbing and Lu, Rongxing},
  journal={IEEE Transactions on Information Forensics and Security},
  year={2024},
  publisher={IEEE}
}

@inproceedings{xie2026healsplit,
  title={Healsplit: Towards self-healing through adversarial distillation in split federated learning},
  author={Xie, Yuhan and Lyu, Chen},
  booktitle={Proceedings of the AAAI Conference on Artificial Intelligence},
  volume={40},
  number={42},
  pages={35931--35939},
  year={2026}
}

@inproceedings{xie2026besplit,
  title={BESplit: Bias-Compensated Split Federated Learning with Evidential Aggregation},
  author={Xie, Yuhan and Lyu, Chen and Huang, Jingrong},
  booktitle={Forty-third International Conference on Machine Learning},
  year={2026}
}

@inproceedings{thapa2022splitfed,
  title={Splitfed: When federated learning meets split learning},
  author={Thapa, Chandra and Arachchige, Pathum Chamikara Mahawaga and Camtepe, Seyit and Sun, Lichao},
  booktitle={Proceedings of the AAAI conference on artificial intelligence},
  volume={36},
  number={8},
  pages={8485--8493},
  year={2022}
}

@misc{lecun1998mnist,
  author       = {Yann LeCun and Corinna Cortes and Christopher J. C. Burges},
  title        = {MNIST Handwritten Digit Database},
  year         = {1998},
  note         = {Available: \url{http://yann.lecun.com/exdb/mnist}}
}

@inproceedings{krauss2023mesas,
  title={Mesas: Poisoning defense for federated learning resilient against adaptive attackers},
  author={Krau{\ss}, Torsten and Dmitrienko, Alexandra},
  booktitle={Proceedings of the 2023 ACM SIGSAC Conference on Computer and Communications Security},
  pages={1526--1540},
  year={2023}
}

@article{blanchard2017machine,
  title={Machine learning with adversaries: Byzantine tolerant gradient descent},
  author={Blanchard, Peva and El Mhamdi, El Mahdi and Guerraoui, Rachid and Stainer, Julien},
  journal={Advances in neural information processing systems},
  volume={30},
  year={2017}
}

@inproceedings{yin2018byzantine,
  title={Byzantine-robust distributed learning: Towards optimal statistical rates},
  author={Yin, Dong and Chen, Yudong and Kannan, Ramchandran and Bartlett, Peter},
  booktitle={International conference on machine learning},
  pages={5650--5659},
  year={2018},
  organization={Pmlr}
}

@inproceedings{guerraoui2018hidden,
  title={The hidden vulnerability of distributed learning in byzantium},
  author={Guerraoui, Rachid and Rouault, S{\'e}bastien and others},
  booktitle={International conference on machine learning},
  pages={3521--3530},
  year={2018},
  organization={PMLR}
}

@article{cao2020fltrust,
  title={Fltrust: Byzantine-robust federated learning via trust bootstrapping},
  author={Cao, Xiaoyu and Fang, Minghong and Liu, Jia and Gong, Neil Zhenqiang},
  journal={arXiv preprint arXiv:2012.13995},
  year={2020}
}

@inproceedings{li2024data,
  title={Data valuation and detections in federated learning},
  author={Li, Wenqian and Fu, Shuran and Zhang, Fengrui and Pang, Yan},
  booktitle={Proceedings of the IEEE/CVF Conference on Computer Vision and Pattern Recognition},
  pages={12027--12036},
  year={2024}
}

@inproceedings{kumar2025fortifying,
  title={Fortifying Federated Learning Towards Trustworthiness via Auditable Data Valuation and Verifiable Client Contribution},
  author={Kumar, K Naveen and Jha, Ranjeet Ranjan and Mohan, C Krishna and Tallamraju, Ravindra Babu},
  booktitle={Proceedings of the Computer Vision and Pattern Recognition Conference},
  pages={4999--5009},
  year={2025}
}

@article{huang2024parameter,
  title={Parameter disparities dissection for backdoor defense in heterogeneous federated learning},
  author={Huang, Wenke and Ye, Mang and Shi, Zekun and Wan, Guancheng and Li, He and Du, Bo},
  journal={Advances in Neural Information Processing Systems},
  volume={37},
  pages={120951--120973},
  year={2024}
}

@inproceedings{xie2024fedredefense,
  title={Fedredefense: Defending against model poisoning attacks for federated learning using model update reconstruction error},
  author={Xie, Yueqi and Fang, Minghong and Gong, Neil Zhenqiang},
  year={2024},
  organization={International Conference on Machine Learning}
}

@article{yuan2025prfl,
  title={PRFL: Personalized and Robust Federated Learning for Non-IID Data with Malicious Participants},
  author={Yuan, Lixiang and Zhang, Jiapeng and Duan, Mingxing and Xiao, Guoqing and Tang, Zhuo and Li, Kenli},
  journal={IEEE Transactions on Mobile Computing},
  year={2025},
  publisher={IEEE}
}

@book{zscore,
  title={Volume 16: how to detect and handle outliers},
  author={Iglewicz, Boris and Hoaglin, David C},
  year={1993},
  publisher={Quality Press}
}

\end{document}